# BERT-GT: Cross-sentence *n*-ary relation extraction with BERT and Graph Transformer

Po-Ting Lai[1] and Zhiyong Lu[1,*]

[1]National Center for Biotechnology Information (NCBI), National Library of Medicine (NLM), National Institutes of Health (NIH), Bethesda, MD, 20894 USA.

*To whom correspondence should be addressed.

Contact: zhiyong.lu@nih.gov

**Abstract:**

A biomedical relation statement is commonly expressed in multiple sentences and consists of many concepts, including gene, disease, chemical, and mutation. To automatically extract information from biomedical literature, existing biomedical text-mining approaches typically formulate the problem as a cross-sentence n-ary relation-extraction task that detects relations among n entities across multiple sentences, and use either a graph neural network (GNN) with long short-term memory (LSTM) or an attention mechanism. Recently, Transformer has been shown to outperform LSTM on many natural language processing (NLP) tasks. In this work, we propose a novel architecture that combines Bidirectional Encoder Representations from Transformers with Graph Transformer (BERT-GT), through integrating a neighbor-attention mechanism into the BERT architecture. Unlike the original Transformer architecture, which utilizes the whole sentence(s) to calculate the attention of the current token, the neighbor-attention mechanism in our method calculates its attention utilizing only its neighbor tokens. Thus, each token can pay attention to its neighbor information with little noise. We show that this is critically important when the text is very long, as in cross-sentence or abstract-level relation-extraction tasks. Our benchmarking results show improvements of 5.44% and 3.89% in accuracy and F1-measure over the state-of-the-art on n-ary and chemical-protein relation datasets, suggesting BERT-GT is a robust approach that is applicable to other biomedical relation extraction tasks or datasets.

## 1 Introduction

The volume of biomedical literature continues to grow rapidly, and it becomes increasingly difficult to maintain and update biomedical knowledge manually. Fortunately, the advance of text-mining techniques is able to greatly help biomedical researchers to retrieve relevant information efficiently in the era of information overload (Fiorini, et al., 2018; Sidhoumi, et al., 2010). In addition, with the rapid growth of the study on personalized medicine and cancer research, it is necessary to extract and search novel genomic variant information from the biomedical literature (Lee, et al., 2020). Biomedical text mining can provide useful data for biomedical curators (Clematide and Rinaldi, 2012) and other researchers in the biomedical communtiy.

Relation extraction (RE) is one of the critical tasks of text mining; it extracts the relation between concepts from the text. It is typically understood as classifying the relation category of a given sentence and an entity pair. Many methods have been proposed for RE. For instance Lee, et al. (2019) and Peng, et al. (2019) adapted the Bidirectional Encoder Representation for Transformer (BERT) architecture (Devlin, et al., 2019) to biomedical literature and clinical records. Their system achieved state-of-the-art performance on several biomedical and clinical benchmarks against many other machine-learning methods.

No matter whether BERT or other relation extraction methods (Li and Ji, 2019; Peng, et al., 2018) are used, they usually focus on classifying a two-entity pair within a single sentence, for example, whether the sentence expresses that the drug "anastrozole" inhibits "breast cancer." Relationships between entities, however, are often expressed across sentences or may involve more than two entities. Two examples are illustrated in Fig. 1. The first example notes that patients with T790M mutations in the epidermal growth factor receptor (EGFR) can be treated with EGFR tyrosine kinase inhibitor (TKI), likes gefitinib. The two sentences collectively express that there is a ternary interaction between the drug (gefitinib), gene/gene product (EGFR tyrosine kinase), and mutation

**Drug-gene-mutation relation example (PMC3594004):**
Three out of five patients with T790M mutations who received gefitinib attained a partial response (all three responders also had exon 19 deletions). Thus, EGFR TKIs are suitable for NSCLC with a T790M mutation that coexists with activating mutations, such as exon 19 deletions or exon 21 point mutation.

**Chemical-disease relation example (PMID:2131034):**
Severe polyneuropathy and motor loss after intrathecal thiotepa combination chemotherapy: description of two cases. Two cases of severe delayed neurologic toxicity related to the administration of intrathecal (IT) combination chemotherapy including thiotepa (TSPA) are presented. Both cases developed axonal neuropathy with motor predominance in the lower extremities 1 and 6 months after IT chemotherapy was administered. Neurologic toxicities have been described with IT-methotrexate, IT-cytosine arabinoside and IT-TSPA. To our knowledge, however, axonal neuropathy following administration of these three agents has not been previously described. In spite of the fact that TSPA is a useful IT agent, its combination with MTX, ara-C and radiotherapy could cause severe neurotoxicity. This unexpected complication indicates the need for further toxicology research on IT-TSPA.

**Fig. 1.** Cross-sentence $n$-ary relation extraction examples.

(T790M). Although, in the second example, the disease "polyneuropathy" and the chemical "IT-cytosine arabinoside (ara-C)" do not co-occur in the same sentence, a relation is asserted between the two entities. Extracting such a relation without gathering information across sentences and multiple entities makes it difficult to predict correctly the relation in both cases.

Over time, the above problem has received the attention of the natural language processing (NLP) field. In 2015, a abstract-level relation extraction dataset was created for an NLP task on chemical-disease relation (CDR) extraction (Wei, et al., 2016). In the dataset, more than 30% of chemical-induced disease pairs are cross-sentences (Xu, et al., 2016). Many cross-sentence methods have been proposed to identify these relations. Verga, et al. (2018) proposed a Bi-affine Relation Attention Network (BRAN), which is designed to extract relations between entities in the biological text without requiring explicit mention-level annotation. Their neural network (NN) architecture, however, is limited in terms of binary entities and cannot be adapted to *n*-ary entities. Peng, et al. (2017) developed a cross-sentence *n*-ary dataset for detecting drug-gene-mutations across sentences. The cross-sentence *n*-ary relation extraction task is understood as detecting relations among *n* entities across multiple sentences. Peng, et al. (2017) further proposed a graph long short-term memory network (graph LSTM) method.

For a cross-sentence *n*-ary relation extraction task, previous methods typically utilize dependency information by incorporating long short-term memory (LSTM) or an attention mechanism into a graph neural network (GNN). In recent year, Vaswani, et al. (2017) propose an encoder-decoder architecture called Transformer, without the use of LSTM or convolutional neural networks (CNN), and they demonstrate that the Transformer outperforms recurrent neural networks (RNNs) with attention to many sequence-to-sequence tasks. Further, Devlin, et al. (2019) used Transformer architecture to develop BERT and proposed two approaches to generate the non-task-specific pre-trained model. They demonstrate that the pre-trained model can be used for different tasks, with transfer-learning approaches, and can outperform state-of-the-art approaches on many NLP tasks.

Because the self-attention mechanism of the Transformer can efficiently utilize the information from the whole input text, we consider that BERT can be used to classify cross-sentence relations as well. We posit, however, that the cross-sentence also may bring noisy information to BERT and can result in difficulties in focusing on explicit information.

In this work, we propose a novel model that adds a Graph Transformer (GT) architecture into BERT (BERT-GT). The Graph Transformer uses a neighbor-attention mechanism, which is a modified version of the self-attention mechanism. In the self-attention mechanism, the whole sentence(s) is used to calculate the attention of the current token. In contrast, the neighbor-attention mechanism calculates its attention based on its neighbor tokens only. Thus, the token can pay attention to its neighbor tokens with limited noise, which is especially important when the text is very long, as in multiple sentences. The main contributions are as follows:

1) In this work, we focus on classifying the biomedical relations of different text lengths, in particular relations across multiple sentences, an issue that is common in biomedical text but not well studied. In addition, both binary and *n*-ary relation classification tasks are considered.

2) While BERT is a robust and state-of-the-art method, its performance is weakened when processing cross-sentence with unrelated relation keywords. In response, we propose a novel architecture that improves BERT by integrating a neighbor-attention mechanism in a Graph Transformer.

3) BERT-GT is evaluated on two independent biomedical benchmarks and our experimental results demonstrate a significant improvement over other state-of-the-art methods on both *n*-ary and CDR datasets. This demonstrates the generalizability and robustness of our method.

## 2  Related Work

This section introduces recent works on the cross-sentence relation extraction benchmark and then gives a brief introduction of a recent graph transformer architecture. Peng, et al. (2017)'s dataset is the most commonly used dataset for the cross-sentence *n*-ary relation extraction. The BioCreative CDR dataset (Wei, et al., 2016) is also widely used to evaluate abstract-level relation extraction. In what follows, we review the recent work on these benchmarks.

### 2.1 Cross-sentence *n*-ary relation extraction

Peng, et al. (2017) proposed a graph LSTM architecture. They first use the NLTK dependency parser to parse text input into a directed dependency graph, and then the graph is divided into two acyclic-directed graphs (left-to-right and right-to-left). Then, they respectively apply two LSTM layers to the two graphs to generate the output sequences. Finally, the prediction is calculated by aggregating the hidden states of the entities with a softmax function. Their architecture achieves an accuracy of 80.7% and outperforms the feature-based classifier (Quirk and Poon, 2017) by an accuracy of 3%.

Song, et al. (2018) then proposed a graph state LSTM (GS LSTM) architecture for the task. They consider that there are two limitations in Peng, et al. (2017)'s architecture. First, Peng, et al. (2017)

can use only the information from a single direction (either left-to-right or right-to-left) in each LSTM layer. Second, Peng, et al. (2017) do not utilize the dependency edge type. Therefore, Song, et al. (2018)'s GS LSTM uses a message-passing mechanism where each token can pass the message to itself in the next LSTM layer, and all hidden states of its dependency words can be passed as well. To address the second problem, they proposed a variation of the LSTM layer, which allows inputs to be weighted according to edge types. GS LSTM achieved an accuracy of 83.2%, which outperforms Peng, et al. (2017)'s architecture by 2.5%.

Most recently, Guo, et al. (2019) introduced an attention-guided graph convolutional networks (AGGCN) model. They consider that the structure of the dependency graph that limits the token can be updated only by its edges. Therefore, they proposed an attention-guided layer to transform the dependency graph, with the word embedding added into the weighted fully-connected matrix, where each cell represents the strength of the edge. They, then proposed a densely connected layer to allow each token to receive a message (hidden state) from all tokens of the previous sub-layer. Finally, Guo, et al. (2019) aggregate the sentence representation and entity representation to predict the relation type. The AGGCN achieved an accuracy of 87.0% and outperformed Song, et al. (2018)'s architecture by 3.8%. However, we also found that Song, et al. (2018)'s open-source implementation has a significant improvement over their original results (see our Experiment section).

**2.2 Abstract-level chemical-induced disease relation extraction**

Xu, et al. (2016) proposed a Support Vector Machine (SVM)-based approach to classify CDR that involves two feature-based classifiers: sentence-level and abstract-level. Their system achieved an $F_1$-measure of 50.73%. Zhou, et al. (2016) proposed an ensemble approach for the CDR task. They developed three classifiers: tree-kernel-based, feature-based, and LSTM-based. Their system achieved an $F_1$-measure of 56%. In addition, they proposed some corpus-specific post-processing rules for boosting performance to 61.31% in $F_1$-measure.

In 2018, Li, et al. (2018) introduced recurrent piecewise convolutional neural networks (RPCNN). In their formulation, a candidate means a unique CDR-ID pair, and an instance means a CDR-NE pair. Therefore, one candidate can have multiple instances. They also proposed that the piecewise convolutional neural networks (PCNN) represent the instances of the same candidate as a means to predict the relation type. The recurrent neural network is used to aggregate the hidden states of the instances from the same candidate to predict the relation type. Their architecture achieved an $F_1$-measure of 59.1%.

Recently, Verga, et al. (2018) proposed Bi-affine Relation Attention Networks (BRANs). They used the Transformer architecture to encode the input text, and the output sequence of the Transformer is passed into two separate multi-layer perceptrons (MLP): head MLP and tail MLP. The first entity representation is selected from the output of the head MLP, and the second entity representation is selected from the output of the tail MLP. Verga, et al. (2018) used a bi-affine function to calculate the correlation between two entity representations and update the hidden sequence. Finally, they aggregate the entity representations of the hidden sequence to calculate the softmax of relation type. Their architecture achieved an $F_1$-measure of 62.1%.

## 2.3 Graph Transformer

Cai and Lam (2020) proposed a graph transformer architecture for the tasks of the abstract meaning representation and the syntax-based machine translation. They proposed a relation-enhanced global attention mechanism that employed gated recurrent unit (GRU) for learning the relation represent of two nodes, then append the mechanism to the self-attention layer. However, their method does not take generalized pre-trained weights learned with large datasets. Instead, we use the original Transformer and our Graph Transformer (GT) simultaneously. By doing so, the original Transformer part can readily use the pre-trained weights, and only pre-trained word embeddings are needed for the GT part. Therefore, our BERT-GT model is easier to adapt to different languages or text compared with Cai and Lam (2020).

## 3 Methods

In this section, we introduce BERT-GT and its detailed implementation. The BERT-GT architecture is illustrated in Fig. 2, and can be seen as having four parts: (1) input representations for two Transformers, (2) Transformer, (3) Graph-Transformer, and (4) layers for aggregating the output states of two Transformers.

The input representations are a tokenized paragraph and its directed graph. The Transformer is the same as BERT's Transformer, and we take it from BERT, which allows BERT-GT to reuse the pre-trained weights from Lee, et al. (2019). GT uses an architecture similar to that of the Transformer but has two modifications. First, the input of GT requires the neighbors' positions for each token. Second, the self-attention mechanism is replaced with a neighbors-attention mechanism, whereby each token's output value is calculated by only its neighbor tokens. Finally, we aggregate the hidden states of two Transformers and use the softmax function to calculate the probability of each label.

Here, we describe the problem formulation for the cross-sentence $n$-ary and the CDR tasks, and then we introduce BERT-GT.

## 3.1 Problem formulation

In our formulation, we assume the text $T$ and entities $E$ are given. Although an input text can be of any arbitrary length, cross-sentence relations are commonly seen within paragraphs instead of across paragraphs or in multiple sections/chapters. Similar to the previous works, we only evaluated BERT-GT on PubMed abstracts. Accordingly, the length of the input text $T$ is within the maximum of an abstract length. Further, an entity can appear more than once in the text and can have more than one ID. Thus, we expand the entity by its IDs. If an entity with many IDs, we will expand the entity into multiple entities, and each has a unique ID. In other words, if an entity of a training/test instance has two IDs, then we will expand it into two training/test instances. Each entity in an instance has a unique ID after the expansion. We follow the problem definition of Guo, et al. (2019). The classification problem is defined as whether a relation $R$ holds for the text $T$ and an entity subset $E'$. In the cross-sentence $n$-ary task, we assume that given a paragraph $T$ that contains a variant/mutation $v$, a gene $g$, and a drug $d$. The classification problem checks whether $R$ holds for the $(d, g, v)$ triple. In the abstract-level chemical-induced disease classification task, a chemical $c$ and a disease $d$ can appear in a abstract multiple times. Each $c$ and $d$ can have one or more chemical ID ($cid$) and disease ID ($did$). The classification task can be defined as checking whether $R$ holds for the $(cid, did)$ pair.

## 3.2 The graph transformer for the BERT

As noted, BERT-GT is illustrated in Fig 2. The Transformer part is adapted from the BERT model (Peng, et al., 2019) and allows us to use the pre-trained model from a large set of biomedical literature. The input of the Transformer is BERT's preprocessed word embeddings (Devlin, et al., 2019) of the text $T$, where $Ei$ denotes the $i$-th entity, and the same entity $Ei$ can appear in $T$ more than once. For example, in Fig. 2, $Ei$ appears twice in the $T$.

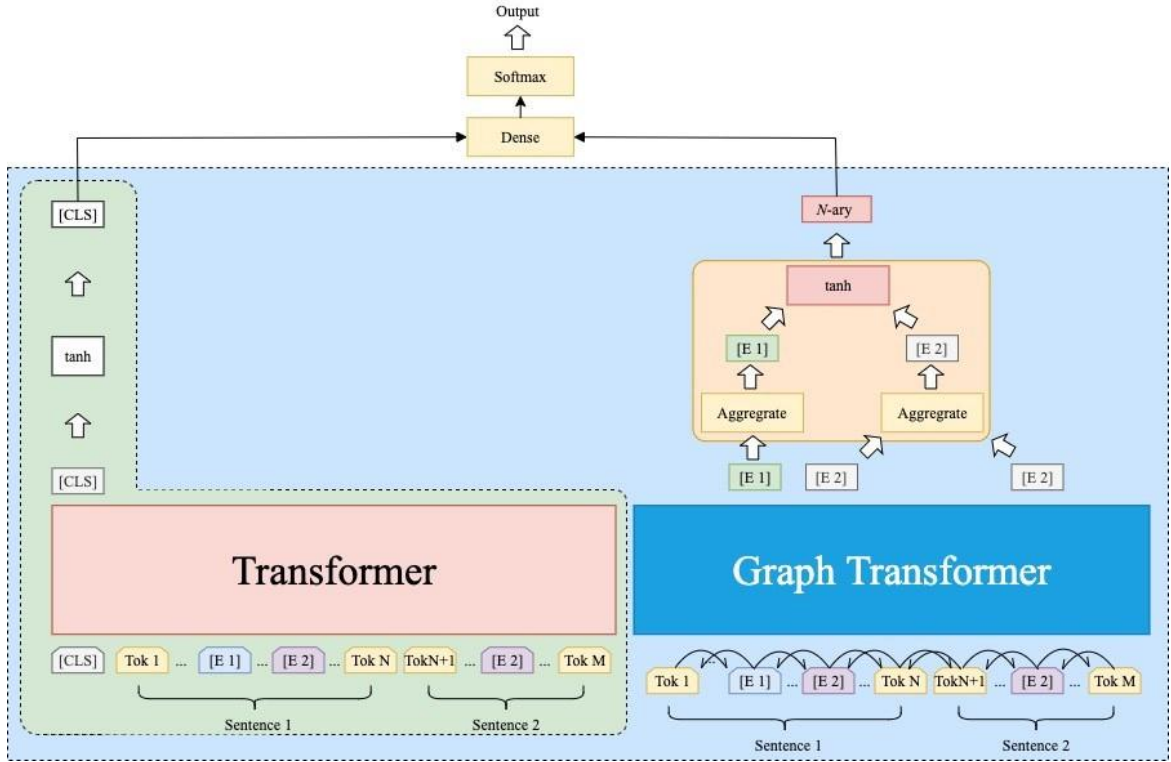

**Fig. 2.** BERT-GT architecture.

The input of GT consists of two representations. One is BERT's preprocessed word embeddings, which is the same as the Transformer part. Another is the neighbors of each token. For the *n*-ary dataset, we use the edges provided by the original dataset, and these edges are generated using the NLTK dependency parser and some heuristic rules. These edges are utilized in many of the previous methods. Because CDR does not provide dependency edges, we use ScispaCy (Neumann, et al., 2019) to parse the paragraph into dependency trees, and the headword and the adjacent two words of each token are treated as neighbors. Notably, many studies (Miwa and Bansal, 2016; Peng, et al., 2018; Xu, et al., 2015) show that the use of the shortest path between entities can improve the relation classification. Therefore, for each entity, we also use the headwords between the entity with the other entities as its neighbors.

The architectures of the Transformer (Vaswani, et al., 2017) and GT are illustrated in Fig. 3. GT replaces the self-attention layer of Transformer with the neighbor-attention layer, and $N_x$ and $N_y$ denote the number of gray block layers.

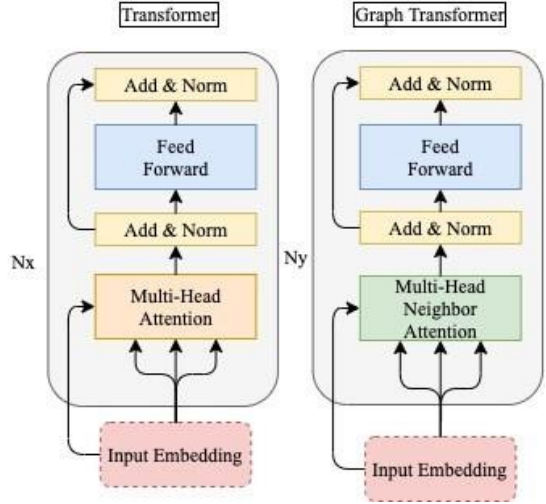

**Fig. 3.** Transformer and Graph Transformer architectures.

In the multi-head self-attention mechanism, each input token representation is divided into $n$ sub-representations of the input token, and $n$ is the number of heads. Assume that the size of the input token representation is $h$, and the size of input token representation for each attention is $h$ divided by $n$. Here we use $h'$ to denote the size of input token $x$ for each attention in Fig. 4, which also illustrates that self-attention is neighbors-attention. Each $x_i$ is transformed into $q_i$, $k_i$, and $v_i$ by multiplying the learnable weighted matrices of $M_q$, $M_k$, and $M_v$. The blue vectors of $x_i$ denote the neighbors of $x_1$. Self-attention calculates $z_1$ by dividing the summation of $q_1$, multiplying by $k_i$ in the text. Although the text length is longer, $z_1$ may suffer from the noisy message. In contrast, we propose a neighbors-attention mechanism, whereby we use only the neighbors of $x_1$ to calculate the neighbors-attention value $a_1$. As illustrated in Fig. 4, the vectors in blue represent the neighbors of $x_i$.

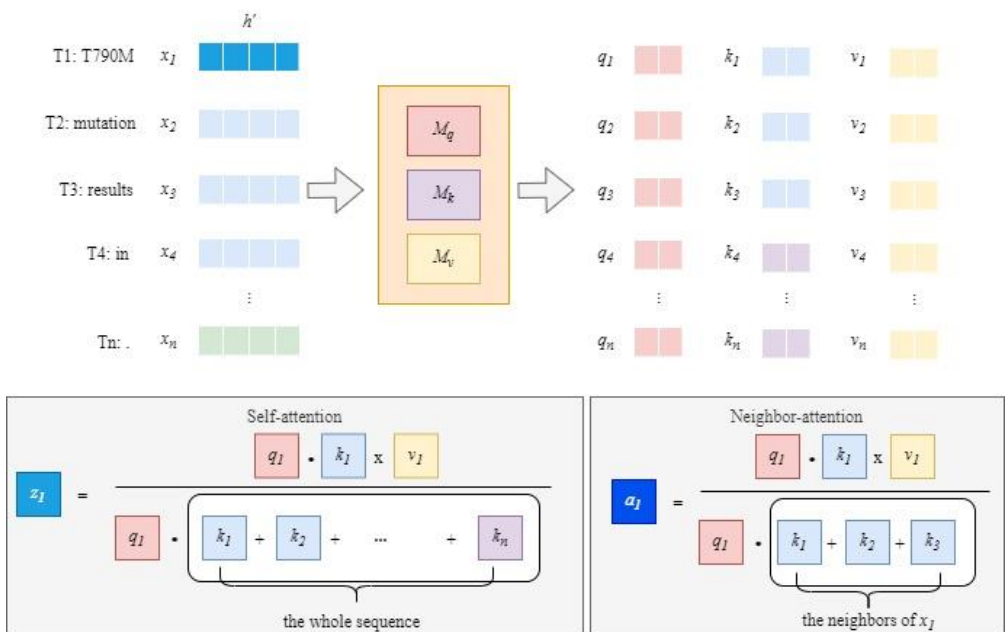

**Fig. 4.** Two different attention mechanisms. For simplification, we use single cells below to represent the vectors above.

**3.3 The output layer of the BERT-GT**

For the classification task, we obtain the sentence representation of the Transformer and the entity representation of GT. The same entity in different positions is aggregated by using the mean average, as illustrated in Fig. 2. We then concatenate the mean averages of different entities to form the entity representation. We also evaluate the effect of using [CLS] for the graph transformer (see our Supplementary Materials). Both sentence representation, for which we choose the output state of the Transformer's first token, and entity representation are passed into the linear transformation layer and follow the dense and softmax layers for the classification. The dimension of the output layer is the same as the number of labels.

## 4 Experimental Results

**4.1 Experimental setup**

To test the generalizability of our aproach, we evaluate our method on two independent tasks: (1) Peng, et al. (2017)'s *n*-ary dataset for the cross-sentence *n*-ary relation extraction task; and (2) Li, et al. (2016) and Wei, et al. (2016)'s CDR dataset for the abstract-level relation classification task. We also evaluate our BERT-GT on traditional sentence-level biomedical relation extraction tasks (see our Supplementary Materials).

For both tasks, we compare with four state-of-the-art models: (1) BERT model (Devlin, et al., 2019); (2) BlueBERT model (Peng, et al., 2019); (3) GS LSTM model (Song, et al., 2018); and (4) AGGCN model (Guo, et al., 2019), through directly using their online source code. Besides, we evaluate the performances of only using graph transformer (GT) architecture. Note that we do not run the state-of-the-art CDR systems on *n*-ary due to two main constraints. First, most CDR systems (Li, et al., 2018; Xu, et al., 2016; Zhou, et al., 2016) are not publicly available for retraining on new datasets. Second, the CDR task contains only the relation between two entities. Therefore the features/architectures of these systems cannot be directly used to classify the relation of *n* entities. In contrast, the *n*-ary systems can be adapted to classify the relation of two entities in abstract-length text.

**4.2 Datasets**

The *n*-ary datasets involve 6,987 drug-gene-mutation (ternary) relation instances and 6,087 drug-mutation (binary) relation instances, and each instance is categorized into one of the five classes: "resistance," "response," "resistance or nonresponse," "sensitivity," and "none." Following Peng, et al. (2017)'s experiment settings, we also considered the two-class classification task, whereby the class "none" is treated as "no" and the remaining classes are treated as "yes." There are, however, two limitations to the *n*-ary datasets. First, although the *n*-ary dataset is cross-sentence, the number of sentences is limited to three per instance. Second, the numbers of instances with "no" and "yes" labels are balanced but are not representative of actual biomedical text. Therefore, we also used the BioCreative CDR (Wei, et al., 2016) dataset for the evaluation.

The BioCreative CDR dataset consists of 1500 PubMed abstracts and is annotated with 4409 chemicals, 5818 diseases, and 3116 chemical-disease relations. The annotations include the spans of entities and their corresponding MeSH IDs. The relations between chemicals and diseases were annotated at the abstract-level, which gives only the chemical-disease MeSH ID pair. In the CDR task, 1000 annotated abstracts are released for participants, and 500 annotated abstracts are used as the test set. Previous work (Xu, et al., 2016) shows that approximately 30% of positive CDR instances are cross-sentence relations. The sizes and comparison of the two datasets are summarized in Table 1.

**Table 1.** Sizes of the $n$-ary and the CDR datasets. "dgm" means "drug-gene-mutation"; "dm" means "drug-mutation"

|  | $n$-ary | CDR |
|---|---|---|
| NE types | drug, gene, mutation | chemical, disease |
| Normalization | None | MeSH ID |
| Relation type | "resistance or nonresponse", "sensitivity", "response", "resistance," and "none" | positive and negative (no relation) |
| Number of instances | resistance or nonresponse: 1479 dgm and 1259 dm instances; sensitivity: 1149 dgm and 1044 dm instances; response: 488 dgm and 501 dm instances; resistance: 292 dgm and 327 instances; none: 3582 dgm and 2956 dm instances; | 3116 positive pairs; 13477 negative pairs |
| Size of dataset set | A total of 6987 dgm and 6087 dm instances for the five-fold cross-validation | training set: 500 abstracts; development set: 500 abstracts; test set: 500 abstracts |

### 4.3 Evaluation metrics

Previous works used different evaluation metrics for the $n$-ary nnd CDR tasks. To compare our method with earlier methods, we follow their evaluation metrics for the two tasks. For the $n$-ary dataset, we use five-fold cross-validation and report the average test accuracy. The partitions of the five-fold cross-validation are the same as those of prior work (Peng et al., 2017). Similar to Song, et al. (2018) and Guo, et al. (2019), we randomly selected 200 instances from the non-test fold as a held-out set for validation and tuning of hyperparameters. For the CDR dataset, we report the precision, recall, and $F_1$-measure on the test set, as the corpus is already divided into the training, development, and test sets. We merged training and development sets and randomly selected 200 abstracts for validation and tuning of hyperparameters. Besides, we evaluate the computational costs and effects of different hyper parameters (see our Supplementary Materials).

### 4.4 Results of the cross-sentence *n*-ary dataset

The performance of the previous state of the art and all comparing methods is shown in Table 2. "Single" means that we report the accuracy of instances within single sentences, while "All" means the accuracy of all instances. It is of note that the released version of the GS LSTM shows better performance than what was reported in Song, et al. (2018). Its performance is also higher than that of AGGCN, especially on the multi-class drug-gene-mutation evaluation metric. Besides, we compare with the pure GT method that removes the Transformer from our BERT-GT. GT scores significantly higher than AGGCN on multi-class, but lower than GS GLSTM. The lower performance might be because GT weights the neighbors by using the neighbor-attention mechanism, and does not utilize the edge type information which is used in GS GLSTM.

Table 2 shows that our BERT-GT method outperforms all of the previous methods. Compared with AGGCN, our method outperforms it by accuracies of 5.07% and 11.73% on two-class and multi-class drug-gene-mutation evaluations, respectively. According to our observation of the error cases that are wrong in the AGGCN but are correct in BERT-GT, the relation information of these cases is less explicit. For example, *"On the other hand, erlotinib$_{DRUG}$ could not inhibit EGFR$_{GENE}$ phosphorylation in H1975 cells because the T790M$_{MUTATION}$ mutation in EGFR causes a*

**Table 2.** Accuracy of different methods on the *n*-ary test set.

| Model | Drug-Mutation | | | Drug-Gene-Mutation | | |
|---|---|---|---|---|---|---|
| | Two-class | | Multi-class | Two-class | | Multi-class |
| | Single | All | All | Single | All | All |
| Logistic regression (Peng, et al., 2017) | 73.90 | 75.20 | - | 74.70 | 77.70 | - |
| Graph LSTM EMBED (Peng, et al., 2017) | 74.30 | 76.50 | - | 76.50 | 80.60 | - |
| Graph LSTM FULL (Peng, et al., 2017) | 75.60 | 76.70 | - | 77.90 | 80.70 | - |
| GS GLSTM (Song, et al., 2018) | 88.15 | 88.56 | 86.92 | 82.86 | 87.60 | 85.91 |
| AGGCN (Guo, et al., 2019) | 85.78 | 85.36 | 76.88 | **86.06** | 86.44 | 79.62 |
| GT | 86.04 | 83.69 | 84.62 | 82.85 | 83.67 | 83.65 |
| BERT (Devlin, et al., 2019) | 88.59 | 91.01 | 90.03 | 83.75 | 90.80 | 90.73 |
| BlueBERT (Peng, et al., 2019) | 89.81 | 92.10 | 92.16 | 85.39 | 91.27 | 91.10 |
| BERT-GT | **91.67** | **93.27** | **93.45** | 85.49 | **91.51** | **91.35** |

*conformation change at the ATP binding pocket, thus decreasing the affinity between erlotinib and EGFR."* The above case was predicted as "None" because both *"not"* and *"inhibit"* indicate a negative relationship. If we take both terms into consideration, however, this indicates a positive relationship: ~26% of the cases contain negation words, *"not"* and *"no."*

In addition, most methods show lower performances on multi-class evaluation. GS GLSTM, BlueBERT, and our BERT-GT method, however, show that their multi-class evaluation can retain the same level of performance as do their two-class evaluations. In addition, our method outperforms BlueBERT on the drug-mutation relation classification by 1.17% and 1.29% on two-class and multi-class, respectively.

**4.5 Results on the CDR dataset**

The performance of our method and recent methods are shown in Table 3. Here, we report only the performance of systems without using domain knowledge in order to compare methods that can be more generalizable. For the same reason, we report their performance that involves no post-processing or ensemble. Table 3 shows that BERT-GT outperforms all of the other methods. In fact, BERT-GT compares favorably to others even when they use additional post-processing methods or ensemble (Gu, et al., 2017; Verga, et al., 2018; Zhou, et al., 2016). These results show that our system has the potential to be adapted to other domains. In addition, our method outperforms

Table 3. Performance on the CDR test set in comparison with state-of-the-art systens

| Model | P | R | F |
|---|---|---|---|
| CD-REST (Xu, et al., 2016) | 59.60 | 44.00 | 50.73 |
| Feature-TreeK-LSTM (Zhou, et al., 2016) | 64.89 | 49.25 | 56.00 |
| + post-processing | 55.56 | 68.39 | 61.31 |
| CNN (Gu, et al., 2017) | 60.90 | 59.50 | 60.20 |
| + post-processing | 55.70 | 68.10 | 61.30 |
| RNN-CNN (Li, et al., 2018) | 55.20 | 63.60 | 59.10 |
| BRAN (Verga, et al., 2018) | 55.60 | 70.8 | 62.10 |
| + ensemble | 63.30 | 67.10 | 65.10 |
| GS LSTM (Song, et al., 2018) | 42.31 | 39.21 | 40.70 |
| AGGCN (Guo, et al., 2019) | 94.23 | 19.46 | 32.26 |
| GT | 30.04 | 74.67 | 42.84 |
| BERT (Devlin, et al., 2019) | 61.41 | 58.82 | 60.09 |
| BlueBERT (Peng, et al., 2019) | 62.80 | 64.45 | 63.61 |
| BERT-GT | **64.94** | **67.07** | **65.99** |

BlueBERT by an $F_1$-measure of 2.38%. Considering the size of the CDR test set is small, we implement a statistically significant test to compare BERT-GT and BlueBERT (see our Supplementary Materials).

GS LSTM and AGGCN, however, show lower recall on the CDR dataset. We observed 25 false-negative cases that are wrong in AGGCN but correct in our proposed method. In ~65% of these cases, the chemical and disease pairs co-occurred in a sentence at least once. AGGCN, however, cannot identify them correctly. Note that the maximum length of the input text is only three sentences in the n-ary dataset, and each instance in the n-ary dataset consists of only one annotated drug-gene-mutation. In contrast, the CDR dataset is in abstract-length, a chemical can be mentioned in different locations of the article. A chemical-disease can have a positive relation at one location in the abstract but have no relation at the rest of the locations. Hence, it is sometimes challenging to use a graph to distinguish the positive location from the negative ones. As a result, unlike our approach BERT-GT, AGGCN and GS LSTM are not robust and generalizable to another domain or a new dataset without additional effort.

## 5    Discussion

**5.1 The performance improvement as a result of adding the graph transformer**

We randomly sampled 50 error cases that were wrong in BERT's predictions but correct in BERT-GT from the *n*-ary test set. The most common error cases of BERT are the instances in which "None" labels were assigned to instances with other labels. These cases constitute approximately 56% of the sampled error cases. For example, *"Interestingly, cells with these mutations also showed greater sensitivity to gefitinib and **erlotinib**$_{DRUG}$ than cells with the **EGFR**$_{GENE}$ mutation (exon 19 deletion), which are associated witsh sensitivity to EGFR inhibitors in NSCLC. Mutations in PI3K (**H1047R**$_{MUTATION}$) have been shown to enhance HER2 mediated transformation by amplifying the ligand induced signaling output of the HER family of RTKs."* The BERT model predicts this as "resistance or non-response." Almost all of these cases are multiple-sentence instances, and it seems that the long instances with some keywords, such as "sensitivity," "associated," and "enhance," which commonly appear in the positive instances, are not used to express the relationship of these entities and can result in the wrong prediction. In addition, we found that 20% of the sampled cases are intra-sentence instances. For example, *"Consistent with this, we found that the introduction of **R1275Q**$_{VARIANT}$ into **EML4-ALK**$_{GENE}$ had no negative impact on sensitivity to **crizotinib**$_{DRUG}$ ( IC50*

*47 ± 8 nm )."* This is misclassified as "resistance" by the BERT model instead of its ground truth "sensitivity." It seems that, although BERT can use the key information from cross-sentence, it also is relatively easier to make an incorrect prediction by using that information and that training on cross-sentence also makes it easier to misclassify short text. In contrast, BERT-GT suffered from fewer of these problems, though a GT-only model does not perform well in our observation. Taken together, this suggests the strengths of GT in either focusing on the neighbor information or predicting short-text instances, when built with the default BERT model.

**5.2 Error analysis, limitations, and future directions**

Here, we discuss the error cases of BERT-GT and future research directions. According to our observation of the BERT-GT error cases on the CDR dataset, in 61% of the error cases, either chemical or disease mentions occurred only once. For example, in the PMID:25986755, the chemical "caffeine" appears only in the first two sentences of the abstract, and the disease "dysplasia" appears only once in the last sentence of the abstract. There is, however, a positive relationship between "caffeine" and "dysplasia" but no explicit information that mentions this relationship; thus, it was misclassified as no relation. Among these error cases, 75% do not have any single sentence that mentions both the target chemical and disease. Therefore, if the entity appears only once in the entire abstract and does not provide explicit information, BERT-GT is likely to generate the wrong predictions.

We also observed that there are some limitations to the *n*-ary dataset. For example, the *n*-ary dataset is a balanced dataset in which the numbers of positive and negative instances are balanced, which is not common in biomedical domains, such as CDR. Further, the sentence length of their instances is no more than three, but, in the biomedical domain, the sentence length of the text is usually longer. Therefore, we consider it a research direction to develop a more representative benchmark for evaluating *n*-ary RE tasks in the future.

**6   Conclusion**

In the biomedical literature, a biomedical relation usually consists of multiple entities and is represented in multiple sentences. The limitations on the architectures/features of previous methods are that they can perform well only on either cross-sentence *n*-ary relation extraction or CDR extraction. In this work, we propose a BERT-GT method. We show that BERT can be used to classify a cross-sentence relation, such as a PubMed abstract, because the attention mechanism makes it able to utilize the key information from the whole text and thus make the prediction. Our GT allows the BERT to use the graph information, which provides the neighbors of each token. The neighbors can be specified to focus on specific information for each token. The results demonstrate that BERT-GT achieves the highest performance on both problems and demonstrates its potential to be applied to a more generalizable relation classification problem.

**Acknowledgements**

We are grateful to the authors of BERT, BlueBERT, BioBERT, GS LSTM, and AGGCN who make their data and source code publicly available. We would like to thank Dr. Zhijiang Guo for helping us to reproduce the results of AGGCN on the *n*-ary dataset. We thank Dr. Chih-Hsuan Wei for his assistance on revising the manuscript.

**Funding**

This research was supported by NIH Intramural Research Program, National Library of Medicine.

*Conflict of Interest:* none declared.

# BERT-GT: Cross-sentence *n*-ary relation extraction with BERT and Graph Transformer (Supplementary Materials)

**A. BERT-GT on Drug-Drug Interaction and ChemProt**

We conducted new experiments to evaluate the BERT-GT and BlueBERT on the drug-drug interaction (DDI) [1] and chemical-protein interaction (ChemProt) [2] datasets. Table 1 shows the performances of BERT-GT, and its performances are slightly higher than the BlueBERT on ChemPort but lower on DDI. Comparing with Peng et.al. [3] and Zhang et.al. [4], the results demonstrate the robustness of our method on traditional tasks.

Table 1. The performances of BERT-GT and state-of-the-art on the same-sentence tasks.

|  | DDI | ChemProt |
|---|---|---|
|  | F | F |
| Zhang et.al. | 72.9 | - |
| Peng et.al. | - | 64.1 |
| BlueBERT | 79.9 | 74.4 |
| BERT-GT | 77.87 | 74.58 |

[1] María Herrero-Zazo, Isabel Segura-Bedmar, Paloma Martínez, Thierry Declerck. The DDI corpus: An annotated corpus with pharmacological substances and drug–drug interactions. Journal of biomedical informatics, 46(5), 914-920. 2013.

[2] Martin Krallinger, Obdulia Rabal, Saber A. Akhondi, Martín Pérez Pérez, Jesús Santamaría, Gael Pérez Rodríguez, Georgios Tsatsaronis, Ander Intxaurrondo, José Antonio López, Umesh Nandal, Erin Van Buel, Akileshwari Chandrasekhar, Marleen Rodenburg, Astrid Laegreid, Marius Doornenbal, Julen Oyarzabal, Analia Lourenço, Alfonso Valencia. Overview of the BioCreative VI chemical-protein interaction track. In Proceedings of BioCreative, pages 141–146. 2017.

[3] Yifan Peng, Anthony Rios, Ramakanth Kavuluru, and Zhiyong Lu. Extracting chemical-protein relations with ensembles of SVM and deep learning models. Database: the journal of biological databases and curation. 2018.

[4] Min-Ling Zhang and Zhi-Hua Zhou. A review on multi-label learning algorithms. IEEE Transactions on Knowledge and Data Engineering, 26(8):1819–1837. 2014.

**B. *t*-test on CDR dataset and 5-fold cross-validation on nary dataset**

We conducted a statistical significance *t*-test on the CDR dataset and compared BERT-GT with BlueBERT, because its size is smaller than the size of n-ary dataset. We merged the CDR training, development, and test sets, and then randomly divide it into 10 different train-test partitions where the sizes of the training and test sets are the numbers of 800 and 500 abstracts, respectively. Their performances are shown in Table 2, and the *p*-value is 0.044 (<0.05), which indicates that BERT-GT statistically outperformed BlueBERT. Table 3 reports the values of the standard deviation on nary dataset.

Table 2. The *t*-test results of BERT-GT and BlueBERT on the CDR dataset.

|          | P     | R     | F     | Δ F  |
|----------|-------|-------|-------|------|
| BlueBERT | 65.65 | 62.27 | 63.89 | 1.47 |
| BERT-GT  | 60.29 | 70.58 | 64.92 | 1.05 |

Table 3. Accuracy of different methods on the *n*-ary test set.

| Model | Drug-Mutation | | | | | | Drug-Gene-Mutation | | | | | |
|---|---|---|---|---|---|---|---|---|---|---|---|---|
| | Two-class | | | | Multi-class | | Two-class | | | | Multi-class | |
| | Single | Δ | All | Δ | All | Δ | Single | Δ | All | Δ | All | Δ |
| Logistic regression (Peng, et al., 2017) | 73.90 | - | 75.20 | - | - | - | 74.70 | - | 77.70 | - | - | - |
| Graph LSTM EMBED (Peng, et al., 2017) | 74.30 | - | 76.50 | - | - | - | 76.50 | - | 80.60 | - | - | - |
| Graph LSTM FULL (Peng, et al., 2017) | 75.60 | - | 76.70 | - | - | - | 77.90 | - | 80.70 | - | - | - |
| GS GLSTM (Song, et al., 2018) | 88.15 | 3.81 | 88.56 | 3.34 | 86.92 | 3.43 | 82.86 | 6.39 | 87.60 | 5.54 | 85.91 | 5.90 |
| AGGCN (Guo, et al., 2019) | 85.78 | 3.13 | 85.36 | 3.31 | 76.88 | 4.29 | **86.06** | 6.81 | 86.44 | 5.98 | 79.62 | 7.20 |
| GT | 86.04 | 4.33 | 83.69 | 4.73 | 84.62 | 4.01 | 82.85 | 6.18 | 83.67 | 5.64 | 83.65 | 5.26 |
| BERT (Devlin, et al., 2019) | 88.59 | 4.01 | 91.01 | 2.62 | 90.03 | 3.83 | 83.75 | 7.82 | 90.80 | 5.58 | 90.73 | 5.75 |
| BlueBERT (Peng, et al., 2019) | 89.81 | 3.12 | 92.10 | 1.42 | 92.16 | 2.01 | 85.39 | 5.10 | 91.27 | 6.27 | 91.10 | 5.40 |
| BERT-GT | **91.67** | 2.70 | **93.27** | 1.40 | **93.45** | 1.57 | 85.49 | 6.90 | **91.51** | 5.60 | **91.35** | 3.41 |

## C. BERT-GT with [CLS]

Because previous works of n-ary relation extraction used the entity representation to aggregate the different context information from different entities. Therefore, we followed their formulation to use the entity representation for GT. To address your point, we have added the [CLS] token mechanism to BERT-GT, and used it as the sentence representation of GT. Table 4 shows it result on the CDR dataset. The performances of using [CLS] for GT are lower than using the entity representation.

Table 4. The performances of BERT-GT with [CLS] on the CDR dataset.

|                    | P     | R     | F     |
| ------------------ | ----- | ----- | ----- |
| BERT-GT            | 64.94 | 67.07 | 65.99 |
| BERT-GT with [CLS] | 59.64 | 69.32 | 64.12 |

### D. Computational Cost

We compared the computational costs of the proposed BERT-GT method against BlueBERT. The hyperparameters of the two models are the same with the sequence length of 512 and the number epoch of 5. Because the BERT-GT model requires additional memory to compute the graph transformer's parameters, thus its running time is three times longer than that of BlueBERT, as shown in Table 5.

Table 5. The computational costs of the BERT and BERT-GT on the CDR dataset.

|  | Training time per epoch |
|---|---|
| BlueBERT | ~20m |
| BERT-GT | ~1 hr. |

## E. Effects of Hyper Parameters

We conducted a new experiment to evaluate the sensitivity of performance on different hyper parameter settings. The results of new experiments are illustrated in the following Figure 1 and 2.

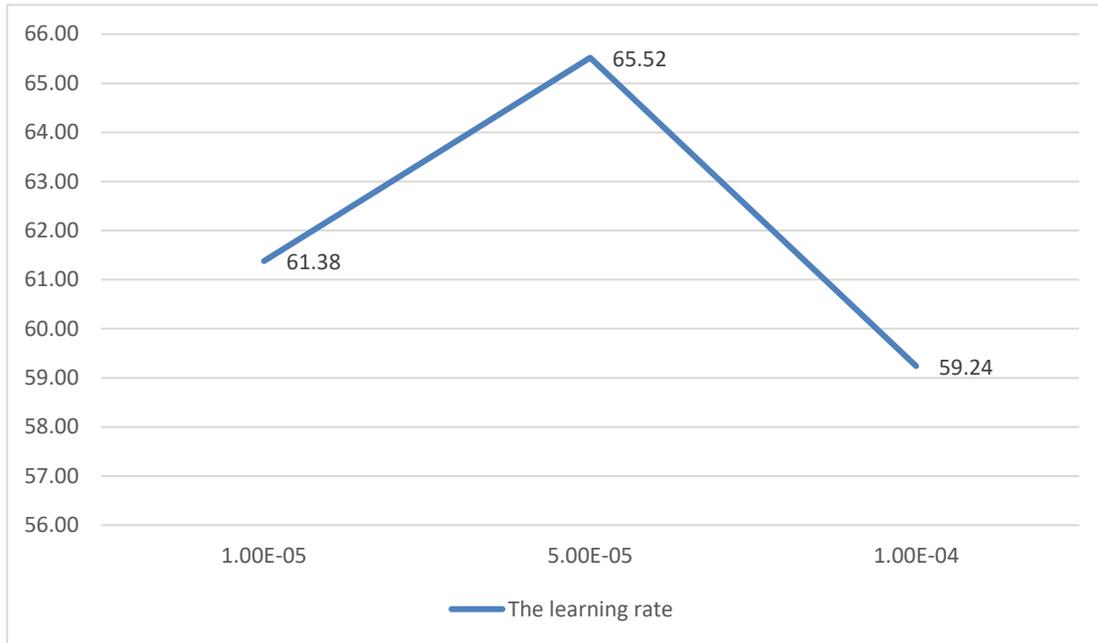

Figure 1. The effects of different learning rate on performances

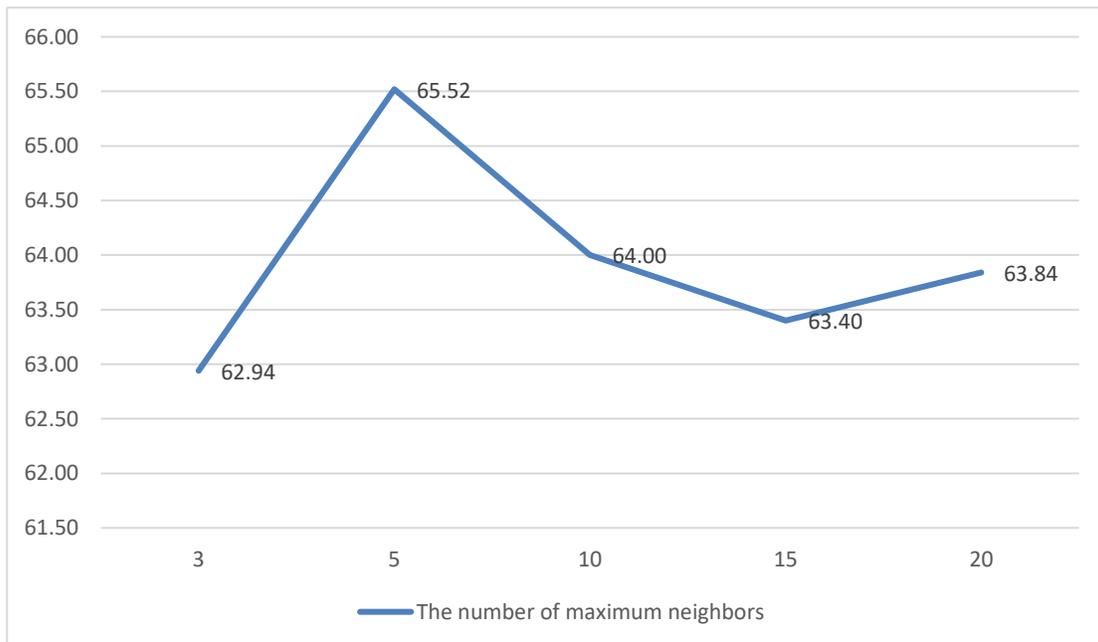

Figure 2. The effects of maximum number of each token's neighbors (neighbor tokens) on performances